\documentclass[10pt,twocolumn,letterpaper]{article}

\usepackage{cvpr}      %

\usepackage{graphicx}
\usepackage{amsmath}
\usepackage{amssymb}
\usepackage{booktabs}
\usepackage{multirow}
\usepackage{cellspace, tabularx}
\usepackage[dvipsnames]{xcolor}
\usepackage{bbm}
\usepackage[accsupp]{axessibility} %

\usepackage[pagebackref,breaklinks,colorlinks]{hyperref}

\usepackage[capitalize]{cleveref}
\crefname{section}{Sec.}{Secs.}
\Crefname{section}{Section}{Sections}
\Crefname{table}{Table}{Tables}
\crefname{table}{Tab.}{Tabs.}

\newcommand{\authorskip}{\hspace{2.5mm}}

\def\cvprPaperID{3716} %
\def\confName{CVPR}
\def\confYear{2022}

\begin{document}

\title{Ditto: Building Digital Twins of Articulated Objects from Interaction}

\author{Zhenyu Jiang \authorskip Cheng-Chun Hsu \authorskip Yuke Zhu \\[2mm]
 Department of Computer Science, The University of Texas at Austin \\{\tt\small \hspace{0mm}\{zhenyu,hsucc,yukez\}@cs.utexas.edu}} 
\maketitle

\begin{abstract}
Digitizing physical objects into the virtual world has the potential to unlock new research and applications in embodied AI and mixed reality. This work focuses on recreating interactive digital twins of real-world articulated objects, which can be directly imported into virtual environments. We introduce Ditto to learn articulation model estimation and 3D geometry reconstruction of an articulated object through interactive perception. Given a pair of visual observations of an articulated object before and after interaction, Ditto reconstructs part-level geometry and estimates the articulation model of the object. We employ implicit neural representations for joint geometry and articulation modeling. Our experiments show that Ditto effectively builds digital twins of articulated objects in a category-agnostic way. We also apply Ditto to real-world objects and deploy the recreated digital twins in physical simulation. Code and additional results are available at \href{https://ut-austin-rpl.github.io/Ditto}{\textcolor{Blue}{\url{https://ut-austin-rpl.github.io/Ditto/}}}
\end{abstract}

\section{Introduction}
\label{sec:intro}

Synthetic data has played a steadily more vital role in fueling emerging AI applications, from training and prototyping computer vision models~\cite{kar2019meta,roberts2021hypersim} to teaching robots to perform physical tasks~\cite{andrychowicz2020learning,mahler2017dex,zhu2017target}. As modern AI models become larger and increasingly data-hungry, virtual platforms and synthetic datasets supply a massive amount of cheap training data. For vision models to benefit from synthetic data, realism is key --- the distribution mismatch between the real and virtual worlds hinders the generalization of models trained in simulation. A promising path towards closing the reality gap is digitizing physical objects and recreating them in virtual environments. Research on 3D vision and SLAM~\cite{avetisyan2019scan2cad,newcombe2011kinectfusion,newcombe2015dynamicfusion,geiger2011stereoscan,wu2017marrnet} has made significant advances in capturing realistic objects and scenes with static 3D models. Nonetheless, the burgeoning body of embodied AI and mixed reality research calls for {\em interactive digital twins} of physical objects that can be spawned in simulated environments and interact with virtual agents. Building digital twins of \textit{articulated objects} is particularly challenging as it requires not only a good understanding of its overall geometry but the part compositions as well as the kinematic relations between the parts.

\begin{figure}
    \centering
    \includegraphics[width=\linewidth]{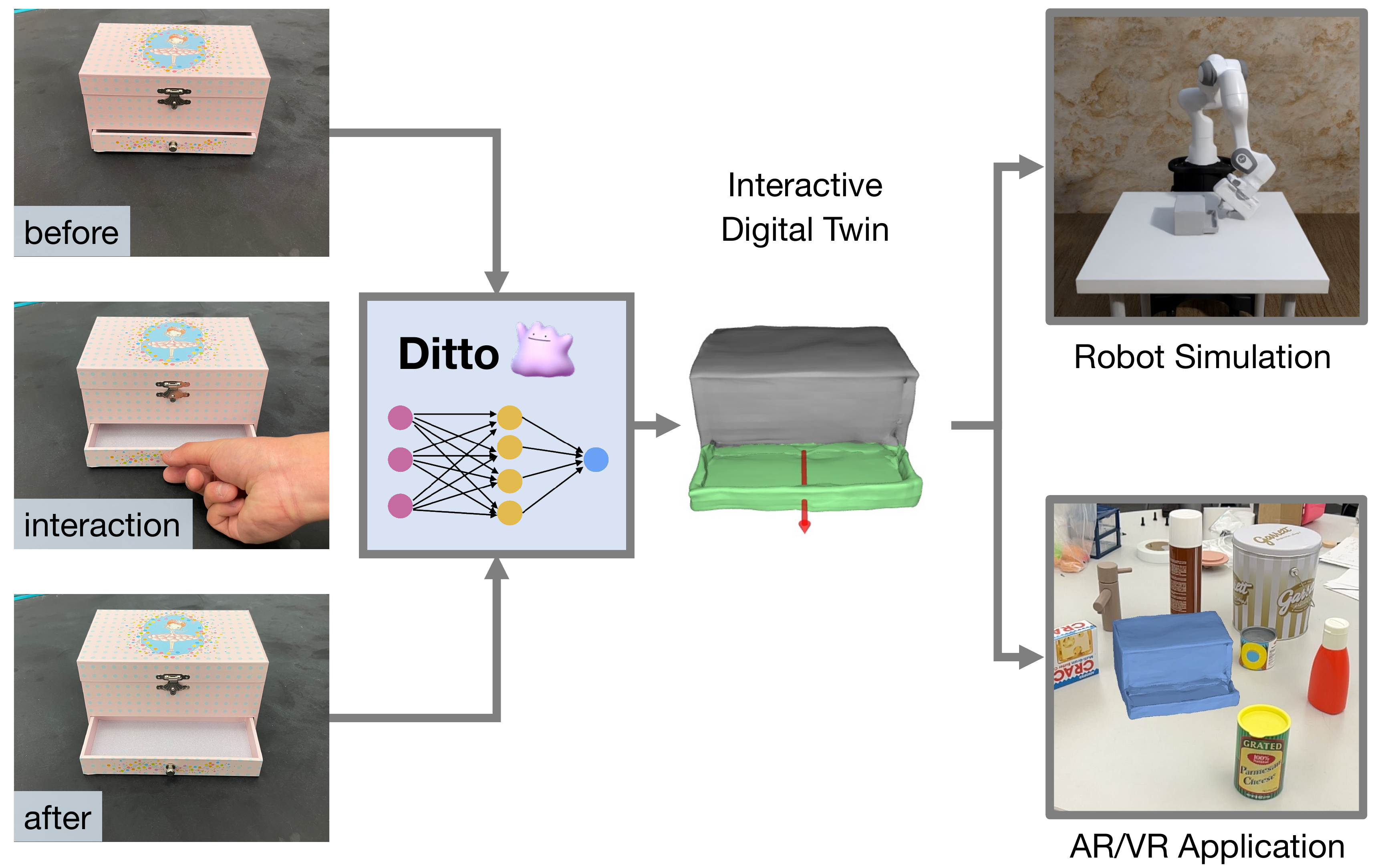}
    \vspace{-2mm}
    \caption{We build digital twins of articulated objects through interactive perception. Given visual observations before and after interaction, our method jointly reconstructs the part-level geometry and articulation model of the object. Our recreated digital twins can be spawned in physics engines and are fully interactive in robot simulation and AR/VR applications.}
    \label{fig:pull}
    \vspace{-2mm}
\end{figure}

Recent efforts in embodied AI platforms~\cite{li2021igibson,kolve2017ai2,szot2021habitat} have incorporated interactive articulated objects, such as cabinets and drawers, in simulated household environments and employed them for training virtual agents. Even so, they heavily rely on graphics designers and engineers to author and curate the object models, limiting the scalability of the asset acquisition process. Developing vision-based methods to automate the estimation~\cite{jain2020screwnet,abbatematteo2019learning} and reconstruction~\cite{mu2021sdf} of articulated objects has been an active line of research, accelerated by new tools developed from the 3D vision community, including geometric deep learning~\cite{kanazawa2018end,omran2018neural,pavlakos2018learning} and implicit neural representations~\cite{deng2020nasa,noguchi2021neural}. The majority of prior work focuses on solving individual components of the problem rather than constructing a full-fledged model. Several recent works~\cite{li2020category,wang2019shape2motion} have studied the joint learning of part segmentation and joint estimation. However, they infer part-level geometry on the point cloud which cannot be used for physical simulation, because physical simulation requires compact geometry of the object such as the mesh for collision computation.

Departing from prior work, we seek the full-fledged virtual recreation of articulated physical objects from unknown categories. These digital twins of articulated objects represent the geometry and physics of individual object parts and their articulation relations (\eg,~prismatic or revolute joints). Category-agnostic articulation estimation from a single image is inherently ambiguous. The parts may move along a prismatic axis or rotate around a revolute axis depending on the underlying kinematic joints. Following pioneer work on the interactive perception of articulated objects~\cite{hausman2015active,martin2014online}, we propose to infer the digital twins from visual observations collected before and after articulated motions (see \cref{fig:pull}). This task comprises three intimately connected challenges: object part segmentation based on motion cues, part reconstruction from a partial point cloud, and articulation estimation of unknown joint types.

We introduce \textbf{Ditto} (\textbf{Di}gital \textbf{t}win of ar\textbf{t}iculated \textbf{o}bjects), an implicit neural representation-based model that jointly predicts part-level geometry and kinematic articulation between the parts. We employ implicit neural representations~\cite{park2019deepsdf,mescheder2019occupancy,chen2019learning,sitzmann2020implicit} to encode continuous and high-resolution 3D information. Ditto is built on top of ConvONets~\cite{peng2020convolutional}, which learns local implicit fields based on convolutional feature grids. The input to Ditto is partial point cloud observations of an articulated object before and after interaction with one of its parts. The key technical challenge is to establish correspondences between these two partial observations.
To achieve this, we encode the point clouds with PointNet++~\cite{qi2017pointnet++} into two sets of subsampled point features. Then we fuse these two sets of point features with a self-attention layer~\cite{vaswani2017attention} and decode the fused subsampled features into dense point features. We construct structured feature grids from the decoded point features. A local feature can be computed from the feature grids at a query 3D coordinate. We learn an implicit occupancy decoder and an implicit segmentation decoder that maps from a 3D coordinate and its local feature to the occupancy/part segmentation label at that coordinate to reconstruct part-level geometry. We use another set of implicit decoders that densely predict the relative joint parameters at each query point to estimate articulated joints. Such dense articulation prediction brings forth more robust articulation estimation than predicting the joint parameters globally. All the implicit decoders can be trained end-to-end.

We evaluate our approach on two datasets of articulated objects~\cite{wang2019shape2motion,abbatematteo2019learning}. Our results demonstrate that Ditto accurately reconstructs part-level geometry and articulation model in a category-agnostic fashion. Ditto achieves superior results across all datasets and metrics compared with the baselines. Furthermore, we apply our method to real-world articulated objects for recreating digital twins. We provide examples of instantiating digital twins in simulation for a virtual robot to interact and transfer the interaction back to the real world.

\section{Related Work}
\label{sec:related_works}

\begin{figure*}[t]
    \centering
    \includegraphics[width=\textwidth]{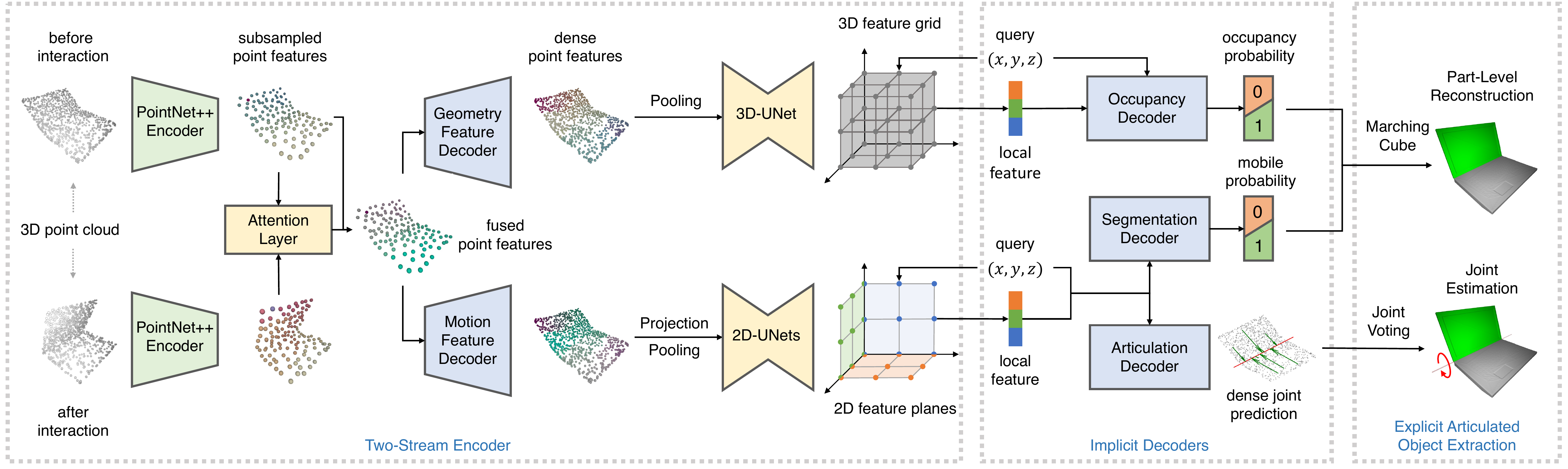}
    \vspace{-5mm}
    \caption{Model architecture of Ditto. The input consists of point cloud observations before and after interaction. After a PointNet++~\cite{qi2017pointnet++} encoder, we fuse the subsampled point features with a simple attention layer. Then we use two independent decoders to propagate the fused point features into two sets of dense point features for geometry reconstruction and articulation estimation separately. We construct feature grid/planes by projecting and pooling the point features and query local features from the constructed feature grid/planes. Conditioning on local features, we use different decoders to predict occupancy, segmentation, and joint parameters with respect to the query points. }
    \label{fig:model}
\end{figure*}

\noindent
\textbf{Articulation Model Estimation.}
Probabilistic methods~\cite{dearden2005learning,sturm2008adaptive,sturm2008unsupervised,sturm2009learning,sturm2011probabilistic} are first used to estimate the articulation relationships between different parts of an articulated object. As articulation can be ambiguous to infer from a single observation, interactive perception methods~\cite{bohg2017interactive,martin2014online,martin2016integrated,hausman2015active,katz2008manipulating,gadre2021act,yi2018deep} have been employed to estimate articulation from action-generated visual stimuli. Conventional methods take a series of sensory observations as input and rely on markers or handcrafted features to track the mobile parts. Recently, deep learning methods have been developed for articulation estimation from raw sensory data~\cite{jain2020screwnet,abbatematteo2019learning,liu2020nothing,jain2021distributional}. Most of these works primarily focus on predicting the articulation parameters. In contrast, our method jointly reconstructs the full 3D geometry and estimates the articulation model.

\vspace{1mm}
\noindent
\textbf{3D Reconstruction of Articulated Objects.}
3D models of articulated objects encode articulation and geometry properties of the objects. Pioneer work from Huang \etal~\cite{huang2012occlusion} uses structure from motion to reconstruct the full point cloud of the object and segment the point cloud using feature-based correspondence. More recently, learning-based methods~\cite{li2020category,wang2019shape2motion} are developed to predict part segmentation together with joint parameters. These works reason about the part-level geometry on the point cloud, which lacks the mesh information required for physical simulation.
A series of methods on reconstructing deformable object~\cite{bozic2021neural,yang2021lasr,yang2021banmo} use articulated bones to represent articulation. These representations loosely constrain the motions of object parts. In contrast, the digital twins require accurate part-level geometry and precise articulation modeling to be simulated in a physics engine.

Closest to our work is A-SDF~\cite{mu2021sdf}, which learns a deep signed function for articulated objects from which a 3D mesh can be extracted. It uses a separate latent code to model the articulation state implicitly. Instead, our method builds full 3D meshes for each part and models their articulations explicitly. The resultant digital twin can spawn in virtual environments for physical interaction.

\vspace{1mm}
\noindent
\textbf{Implicit Neural Representations.}
Our method builds on top of recent work on implicit neural representations~\cite{chen2019learning,mescheder2019occupancy,park2019deepsdf}. These works encode a 3D shape with the iso-surface of an implicit function. These implicit models are parametrized with deep networks so that they are capable of representing complex shapes smoothly and continuously in high resolution. Aiming at better scalability and finer details, several approaches~\cite{peng2020convolutional,liu2020neural,genova2020local} learn local implicit decoders and condition the implicit representations on local features instead of a global shape feature. Specifically, our model extends ConvONets~\cite{peng2020convolutional} with a stronger encoder and a fusing module for processing two input data streams. 

\vspace{1mm}
\noindent
\textbf{Physical Simulation with Articulated Objects.}
Physical simulators have become a vital tool for embodied AI research. A growing trend is shifting from static 3D scenes for visual navigation~\cite{zhu2017target,kolve2017ai2,savva2017minos} to interactive environments that support physical interaction between the robot and the objects~\cite{deitke2020robothor,szot2021habitat,li2021igibson}. Interactive 3D assets are the key elements to construct these simulators. Existing interactive 3D assets are mostly authored and refined by 3D artists~\cite{szot2021habitat,xiang2020sapien,mo2019partnet,wang2019shape2motion} or procedurally generated~\cite{abbatematteo2019learning}. Our method builds interactive digital twins of daily articulated objects directly from visual observations. It has the potential to accelerate the acquisition of realistic interactive 3D assets.

\section{Problem Formulation}
\label{sec:problem}

We study the problem of recreating interactive digital twins of articulated objects from a pair of sensory observations before and after an interaction. Digital twins are commonly represented in standard 3D formats, such as URDF\footnote{\url{http://wiki.ros.org/urdf}}, such that they can be imported into physics engines. To enable physical interaction in the virtual world, a digital twin of an articulated object constitutes a \textit{kinematic tree}, where the nodes define the geometry and physical properties (\eg mass and friction) of individual parts and the edges define the kinematic joints between the parts. This work focuses on estimating part geometry and kinematic articulation while setting the physical properties to default values based on real-world statistics.

Given an articulated object from an unknown category, we interact with the object to change the articulation state. Without the loss of generality, we assume only one part is moved after the interaction, which we call the mobile part. The input to our method is a pair of point cloud observations $\mathcal{P}_1, \mathcal{P}_2 \in \mathbb{R}^{N\times 3}$ of the articulated object before and after an interaction. $N$ is the number of input points. The objective is to segment and reconstruct the 3D geometry for static and mobile parts, estimate the joint parameters that connect these two parts, and relative change of the joint states.

For articulation estimation, we consider the 1D revolute joints and 1D prismatic joints. We follow Li~\etal\cite{li2020category} and parameterize the two types of joints as follows. The parameters of a prismatic joint consist of the direction of the translation axis $\mathbf{u}^p \in \mathbb{R}^3 $ and the joint state $c^p$. The joint state $c^p$ is defined as the relative translation distance between the two observations. The parameters of a revolute joint consist of the direction of the revolute axis $\mathbf{u}^r \in \mathbb{R}^3 $, a pivot point $\mathbf{q} \in \mathbb{R}^3 $ on the revolute axis and the joint state $c^r$. The joint state $c^r$ is defined as the relative rotation angle between the two observations.
\vspace{-0.2cm}
\section{Method}
\label{sec:method}

We now present Ditto, a learning framework that builds digital twins of articulated objects through interactive perception. Ditto jointly learns part-level geometry reconstruction and articulation model estimation with structured feature grids and unified implicit neural representations. \cref{fig:model} illustrates the overall model architecture. Ditto consists of a two-stream encoder that fuses two input point clouds and multiple implicit decoders for geometry and articulation. The model is jointly optimized with a combination of loss functions on geometry reconstruction and articulation estimation. Upon inference, we extract explicit models of articulated objects from the implicit decoders.

\subsection{Two-Stream Encoder}
\label{sec:encoder}

To jointly learn the 3D reconstruction and articulation model estimation, we need to extract features that fuse the information from the input pair of point clouds. We build our encoder based on ConvONets~\cite{peng2020convolutional}, the state-of-the-art implicit representation-based 3D reconstruction method.

We use an attention layer~\cite{vaswani2017attention} to fuse the two sets of point features of two input point clouds. The complexity of attention operations exhibits quadratic growth with respect to the number of points. To process more dense point clouds which capture finer details of the object, we use a PointNet++~\cite{qi2017pointnet++} encoder $\mu_\text{enc}$ to obtain two sets of subsampled point features $f_1 = \mu_\text{enc}(\mathcal{P}_1)$ and $f_2 = \mu_\text{enc}(\mathcal{P}_2)$, where $f_1, f_2 \in \mathbb{R}^{N'\times d_\text{sub}}$, $N' < N$ is the number of the subsampled points, and $d_\text{sub}$ is the dimension of the subsampled point features. A scaled dot-product attention operation is applied to these subsampled feature points
\begin{equation}
    Attn_{12} = \text{softmax}(\frac{f_1 f_2^T}{\sqrt{d_\text{sub}}})f_2, \quad
    f_{12} = [f_1, Attn_{12}].
\end{equation}
The fused subsampled point features $f_{12} \in \mathbb{R}^{N'\times 2d_\text{sub}}$ is the concatenation of the $f_1$ and the output of attention. Then we use two PointNet++ decoder $\nu_\text{geo}$ and $\nu_\text{art}$ to propagate the fused subsampled point features into dense features aligned with the original points
\begin{equation}
    f_\text{geo} = \nu_\text{geo}(f_{12})\quad \text{and} \quad f_\text{art} = \nu_\text{art}(f_{12}),
\end{equation}

\noindent where $f_\text{geo}, f_\text{art} \in \mathbb{R}^{N\times d_\text{dense}}$ are $d_\text{dense}$-dim point features aligned with $\mathcal{P}_1$. We use two separate sets of dense point features because geometry reconstruction mainly exploits the static observation, while the articulation estimation relies more on the correspondence between two observations. Also, these features are processed separately. $f_\text{art}$ is projected into 2D feature planes and $f_\text{geo}$ is projected into voxel grids as in the ConvONets~\cite{peng2020convolutional}. The points that fall into the same pixel cell or voxel cell are aggregated together via max pooling. This projection operation greatly reduces the computation cost while keeping the spatial distribution of feature points. We apply the projection to three canonical planes $\mathbf{c}$ and a coarse voxel grid $\mathbf{v}$. The resulting feature planes and grid are processed with independent 2D and 3D UNets~\cite{ronneberger2015u}. The output voxel feature grid is used for the geometry implicit decoder, and the feature planes are used for the articulation implicit decoders. Geometry reconstruction requires dense feature grids for fine-grained and local reasoning, while sparse feature planes are sufficient for articulation estimation. Therefore we choose this separate feature representation.

\subsection{Implicit Decoders}
\label{sec:decoder}

Motivated by recent works that demonstrate the continuity and versatility of implicit neural representations~\cite{mescheder2019occupancy,park2019deepsdf,jiang2021synergies,florence2021implicit}, we design implicit decoders for both geometry and articulation reasoning. As both tasks require reasoning about fine-grained geometry details, we condition the implicit decoders on local features. These local features can be computed from the feature grid/planes using trilinear/bilinear sampling given a query 3D coordinate $\mathbf{p}\in\mathbb{R}^3$.

\subsubsection{Geometry Implicit Decoder}
Our geometry implicit decoder is a mapping from a coordinate $\mathbf{p} \in \mathbb{R}^3$ to the occupancy probability $o(\mathbf{p})$ at the coordinate. The occupancy $o(\mathbf{p})$ should be 1 if the point $\mathbf{p}$ is occupied by the object and 0 otherwise. We query the local feature $\psi_{\mathbf{p}}^\mathbf{v} $ from the feature grid $\mathbf{v}$ using trilinear sampling.
Conditioned on the query point coordinate $\mathbf{p}$ and local feature $\psi_{\mathbf{p}}^\mathbf{v}$, our geometry implicit decoder predicts the occupancy probability:
\begin{equation}
\begin{aligned}
    f_{\theta_{o}}(\mathbf{p}, \psi_{\mathbf{p}}^\mathbf{v}) \rightarrow o(\mathbf{p}).
\end{aligned}
\end{equation}

\subsubsection{Articulation Implicit Decoders}
Our articulation implicit decoders map from an arbitrary point $\mathbf{p}_\text{in}$ inside the object to the segmentation label and joint parameters with respect to this point. We only consider the space inside the object because articulation is only meaningful for points in this space. We query the local feature $\psi_{\mathbf{p}_\text{in}}^\mathbf{c}$ from the feature planes $\mathbf{c}$ using bilinear sampling.

\vspace{1mm}
\noindent \textbf{Segmentation.}
 Since we assume that only one joint's state is changed due to the interaction, we can segment the object into the static and mobile parts during each interaction. Therefore we predict a binary segmentation label $s(\mathbf{p}_\text{in})$ where 0 stands for the static part and 1 stands for the mobile part. Our segmentation implicit decoder predicts the segmentation probability conditioning on the local feature:
\begin{equation}
\begin{aligned}
    f_{\theta_\text{seg}}(\mathbf{p}_\text{in}, \psi_{\mathbf{p}_\text{in}}^c) \rightarrow s(\mathbf{p}_\text{in}).
\end{aligned}
\end{equation}
{\flushleft \textbf{Joint Parameters.}}
 Even though the joint is a global property of an articulated object, we use a per-point representation to better utilize our structured feature representation and get a more robust estimation through voting. We share the feature planes for articulation and segmentation prediction since the articulation can be inferred from motion cues like segmentation. First, we use an implicit decoder to predict joint type $p_{j_\text{type}}$:
\begin{equation}
\begin{aligned}
     f_{\theta_\text{type}}(\mathbf{p}_\text{in}, \psi_{\mathbf{p}_\text{in}}^c) \rightarrow p_{j_\text{type}}(\mathbf{p}_\text{in}).
\end{aligned}
\end{equation}
Then we use two implicit decoders to predict parameters and states of prismatic joints and revolute joints. The prismatic joint is defined by its translation axis direction, a 3D unit vector $\mathbf{u}^p$. The joint state is the translation distance $c^p$ resulting from the interaction. Revolute joint parameters include the rotation axis direction  $\mathbf{u}^r$. Different from the prismatic joint, the position of the revolute joint axis also matters. We follow Li \etal~\cite{li2020category} and define the joint position with respect to point $\mathbf{p}_\text{in}$ as the projection of $\mathbf{p}_\text{in}$ to the axis, represented by a 3D unit vector for the projection direction $\mathbf{d}_{\mathbf{p}_\text{in}} ^r$ and a scalar $h_{\mathbf{p}_\text{in}}^r$ for the projection distance. The joint state of the revolute joint is the rotation angle $c^r$ resulting from the interaction. We directly predict these parameters with the implicit decoders:
\begin{equation}
\begin{aligned}
     f_{\theta_{\text{param}_p}}(\mathbf{p}_\text{in}, \psi_{\mathbf{p}_\text{in}}^c) &\rightarrow [\mathbf{u}^p, c^p], \\
     f_{\theta_{\text{param}_r}}(\mathbf{p}_\text{in}, \psi_{\mathbf{p}_\text{in}}^c) &\rightarrow [\mathbf{u}^r, \mathbf{d}_{\mathbf{p}_\text{in}} ^r, h_{\mathbf{p}_\text{in}}^r, c^r] .
\end{aligned}
\end{equation}

\subsection{Training}
\label{sec:training}

Our method does not assume known joint types during inference. Therefore, the model can be trained with data from different categories. The loss for training consists of two parts: the geometry loss and the joint loss. The geometry loss optimizes the part-level geometry reconstruction, and the joint estimation loss optimizes joint estimation.

\vspace{2mm}
\noindent
\textbf{Geometry Loss.} We apply standard binary cross-entropy losses on occupancy and segmentation predictions,
\begin{equation}
\begin{aligned}
\mathcal{L}_\text{occ} &= \sum_{\mathbf{p}} BCE(o(\mathbf{p}), \hat{o}(\mathbf{p})), \\
\mathcal{L}_\text{seg} &= \sum_{\mathbf{p}_\text{in}} BCE(s(\mathbf{p}_\text{in}), \hat{s}(\mathbf{p}_\text{in})),
\end{aligned}
\end{equation}
\noindent where $\hat{o}(\mathbf{p})$ is the ground truth occupancy at $\mathbf{p}$ and $\hat{s}(\mathbf{p}_\text{in})$ is the ground truth segmentation label at $\mathbf{p}_\text{in}$.

\vspace{2mm}
\noindent
\textbf{Joint Loss.} 
We have three implicit decoders that predict joint type, prismatic joint parameters, and revolute joint parameters respectively. For joint type prediction, we also apply the standard binary cross entropy loss. The joint type loss is denoted as $\mathcal{L}_\text{type} = \sum_{\mathbf{p}_\text{in}} BCE(p_{j_\text{type}}(\mathbf{p}_\text{in}), \hat{t})$, where $\hat{t}$ is the ground truth joint type.

{\flushleft \textit{Prismatic Joint.}}
We penalize the orientation difference between the estimated joint axis and the ground truth one with loss $\mathcal{L}_{\text{ori}_p} = \text{arccos}( \mathbf{u}^p \cdot \hat{\mathbf{u}}^p)$. The state prediction is optimized with simple $\ell_1$ loss $\mathcal{L}_{\text{state}_p} = |c^p - \hat{c}^p|$, where $\hat{c}^p$ is the ground truth joint state. Besides, we also minimize the different between the predicted displacement and ground truth one. The state prediction and parameter prediction can be jointly optimized with this loss $\mathcal{L}_{\text{disp}_p} = ||c^p \mathbf{u}^p - \hat{c}^p \hat{\mathbf{u}}^p||$. All together we have joint loss of the prismatic joint
\begin{equation}
\begin{aligned}
\mathcal{L}_{\text{param}_p} = \sum_{\mathbf{p}_\text{in}} (\mathcal{L}_{\text{ori}_p} + \mathcal{L}_{\text{state}_p} + \mathcal{L}_{\text{disp}_p}).
\end{aligned}
\end{equation}

{\flushleft \textit{Revolute Joint.}}
The loss for axis orientation and joint state of the revolute joint is the same as the prismatic joint, denoted as $\mathcal{L}_{\text{ori}_r}$ and $\mathcal{L}_{\text{state}_r}$. We apply the same loss for orientation of the projection direction $\mathbf{d}_{\mathbf{p}_\text{in}} ^r$ and projection distance $h_{\mathbf{p}_\text{in}}^r$, which are added together to form $\mathcal{L}_{\text{pos}_r}$. Thanks to our dense joint representation, displacement based loss can be also applied to revolute joint parameters prediction. For each point $\mathbf{p}_\text{in}$, we compute predicted rotation matrix $R_{\mathbf{p}_\text{in}}$ and ground truth one $\hat{R}_{\mathbf{p}_\text{in}}$ based on predicted and ground truth axis orientation and rotation angle. We also locate the estimated pivot point on the axis $\mathbf{q}_{\mathbf{p}_\text{in}} = \mathbf{p}_\text{in} + h_{\mathbf{p}_\text{in}}^r \mathbf{d}_{\mathbf{p}_\text{in}} ^r$. Then the displacement can be computed as $\mathbf{l}_{\mathbf{p}_\text{in}} = R_{\mathbf{p}_\text{in}} (\mathbf{p}_\text{in}-\mathbf{q}_{\mathbf{p}_\text{in}}) + \mathbf{q}_{\mathbf{p}_\text{in}}$. The ground truth displacement $\hat{\mathbf{l}}_{\mathbf{p}_\text{in}}$ can be computed similarly with the ground truth parameters. And the displacement loss $\mathcal{L}_{\text{disp}_r} = ||\mathbf{l}_{\mathbf{p}_\text{in}}  - \hat{\mathbf{l}}_{\mathbf{p}_\text{in}}||$. Moreover, we apply an extra loss on rotation matrix following ScrewNet~\cite{jain2020screwnet} $\mathcal{L}_{\text{rot}_r} = ||\mathbf{I}_{3,3} - R_{\mathbf{p}_\text{in}}\hat{R}_{\mathbf{p}_\text{in}}^T||$. All together we have joint loss of the revolute joint
\begin{equation}
\begin{aligned}
\mathcal{L}_{\text{param}_r} = \sum_{\mathbf{p}_\text{in}} (\mathcal{L}_{\text{ori}_r} + \mathcal{L}_{\text{state}_r} + \mathcal{L}_{\text{pos}_r} + \mathcal{L}_{\text{disp}_r} + \mathcal{L}_{\text{rot}_r}).
\end{aligned}
\end{equation}
Since the joint type is unknown, we need to dynamically apply the prismatic or revolute joint loss based on the joint type. The full loss is
\begin{equation}
\begin{aligned}
\mathcal{L} = \mathcal{L}_\text{occ} + \mathcal{L}_\text{seg} + \mathcal{L}_\text{type} + \mathbbm{1}_{p} \mathcal{L}_{\text{param}_p} + \mathbbm{1}_{r} \mathcal{L}_{\text{param}_r},
\end{aligned}
\end{equation}

\noindent where $\mathbb{I}_{p}$ and $\mathbb{I}_{r}$ are indicators of whether the ground truth joint type is prismatic or revolute.

\begin{table*}[t]
    \small
    \centering
    \begin{tabular}{l|l|cc|c|cc}
    \toprule
\raisebox{\dimexpr-1\normalbaselineskip-2\cmidrulewidth-2\aboverulesep}[0pt][0pt]{Dataset} & \raisebox{\dimexpr-1\normalbaselineskip-2\cmidrulewidth-2\aboverulesep}[0pt][0pt]{Method} & \multicolumn{2}{c|}{Geometry} & \multicolumn{3}{c}{Joint} \\
      \cmidrule{3-7}
      & & \raisebox{\dimexpr-1\normalbaselineskip-1\cmidrulewidth-1\aboverulesep}[0pt][0pt] {\shortstack[c]{Whole\\Chamfer Distance $\downarrow$}}  & \raisebox{\dimexpr-1\normalbaselineskip-1\cmidrulewidth-1\aboverulesep}[0pt][0pt] {\shortstack[c]{Mobile\\Chamfer Distance $\downarrow$}}  & Prismatic & \multicolumn{2}{c}{Revoulute} \\
      \cmidrule{5-7}
      & & & & Angle Err $\downarrow$ & Angle Err $\downarrow$ & Pos Err $\downarrow$ \\
     \midrule
     \multirow{4}{*}{\shortstack[l]{Synthetic \\ Dataset~\cite{abbatematteo2019learning}}} & A-SDF~\cite{mu2021sdf} & 2.48 & - & - & - & - \\
     & Correspondence~\cite{choy2019fully} & 2.13 & 93.2 & 10.3 & 46.5 & 0.46 \\
     & Global Joint~\cite{jain2020screwnet} & 0.54 & 37.7 & 0.69 & 52.0 & 0.13 \\
     \cmidrule{2-7}
     & Ditto (Ours) & \textbf{0.38} & \textbf{0.21} & \textbf{0.06} & \textbf{0.72} & \textbf{0.03} \\
     \midrule
     \multirow{6}{*}{\shortstack[l]{Shape2Motion\\\hspace{-0.006in}\cite{wang2019shape2motion}}} & Correspondence~\cite{choy2019fully}  & 2.22 & 35.7 & 15.2 & 45.5 & 0.28 \\
     & Global Joint~\cite{jain2020screwnet} & 0.90 & 64.9 & 1.36 & 79.8 & 0.17\\
     \cmidrule{2-7}
      & Share Feature & 0.75 & 10.7 & \textbf{0.07} & 2.22 & 0.04 \\
     & Concat Fusion & 0.97 & 3.09 & 0.17 & 3.13 & 0.03 \\
     & Share Decoder & \textbf{0.68} & 3.30 & 0.19 & 1.93 & \textbf{0.02} \\
     \cmidrule{2-7}
     & Ditto (Ours)& 0.72 & \textbf{0.42} & 0.08 & \textbf{1.36} & \textbf{0.02} \\
     \bottomrule
    \end{tabular}
    \vspace{-3mm}
    \caption{Quantitative results of geometry reconstruction and articulation estimation on Shape2Motion~\cite{wang2019shape2motion} and synthetic~\cite{abbatematteo2019learning} datasets.}
    \vspace{-4mm}
    \label{tab:ablation}
\end{table*}

\subsection{Explicit Articulated Object Extraction}

We need to extract the explicit part-level meshes and articulation model from the learned implicit representation to build interactive digital twins that can be spawned in a virtual environment.

\vspace{1mm}
\noindent
\textbf{Part-Level Mesh Extraction.}
To reconstruct part-level meshes, we mask the occupancy query results with segmentation query results. The occupancy query results of mobile part and static part are 
\begin{equation}
\begin{aligned}
o_{m}(\mathbf{p}) &= \mathbbm{1}[o(\mathbf{p}) > t_\text{occ}] \mathbbm{1}[s (\mathbf{p}) > t_\text{seg}], \\
o_{s}(\mathbf{p}) &= \mathbbm{1}[o(\mathbf{p}) > t_\text{occ}] \mathbbm{1}[s (\mathbf{p}) \leq t_\text{seg}],
\end{aligned}
\end{equation}

\noindent where $t_\text{occ}$ and $t_\text{seg}$ are thresholds for the predicted occupancy and segmentation probability. Then we apply Multiresolution IsoSurface Extraction~\cite{mescheder2019occupancy} and Marching Cube~\cite{lorensen1987marching} to extract per-part surface meshes.

\vspace{2mm}
\noindent
\textbf{Global Articulation Model Extraction.}
We use a simple average voting strategy to aggregate the dense joint prediction. During the mesh extraction, we can sample numerous points inside the mesh with their predicted label. Because the object's motion determines the articulation model, we only let points inside the mobile part vote for the global joint. For joint axis direction and joint state of both types of joint, we average all mobile points' predictions. As for the position of the revolute axis, we compute the pivot point coordinate of each mobile point using the predicted projection direction and distance. Then we average the results of all mobile points and get the estimated pivot point on the axis.

\section{Experiments}
\label{sec:exp}

We examine Ditto's ability to recreate digital twins of articulated objects. We first perform systematic quantitative evaluations on two 3D asset datasets, showing that Ditto can accurately reconstruct the geometry and estimate the articulation model. We then qualitatively show that our method generalizes well to objects in the real world.

\subsection{Datasets}

We conduct experiments on two 3D articulated object datasets, the synthetic objects dataset provided by Abbatematteo \etal~\cite{abbatematteo2019learning} and the Shape2Motion dataset~\cite{wang2019shape2motion}. The synthetic dataset contains procedurally generated articulated objects. Shape2Motion contains human-designed objects. We select four categories from each dataset. For Shape2Motion dataset, we choose four categories with more than 30 instances. We choose 4 out of 6 categories for the synthetic dataset because the other two are very similar to the chosen ones. During data generation, we randomly spawn an object in simulation and set the object into random start and end states to mimic articulated motions. In each state, we fuse multi-view depth images into point cloud observation. Even though we use multi-view depth images, the point cloud may still be incomplete due to the self-occlusion of the objects. We generate occupancy data points for each part separately for ground truth geometry and aggregate the samples to get shape-level occupancy and segmentation. Ground-truth articulation is directly acquired from the simulator and the ground-truth occupancy is queried from the ground truth mesh as in \cite{peng2020convolutional}.

\begin{figure*}[t]
    \centering
    \includegraphics[width=0.95\textwidth]{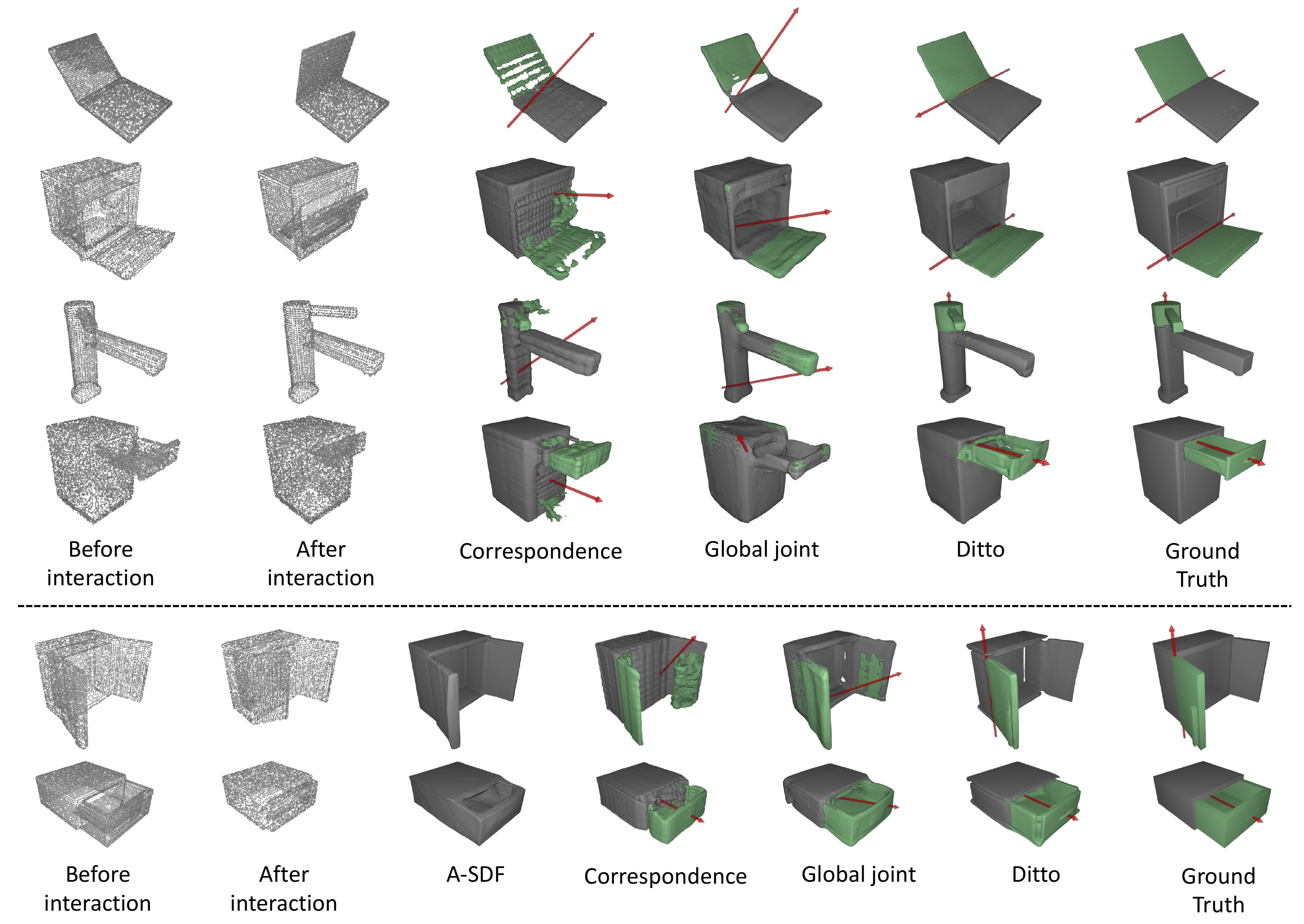}
    \vspace{-2mm}
    \caption{Reconstructed unseen articulated objects in Shape2Motion~\cite{wang2019shape2motion} (top) and synthetic~\cite{abbatematteo2019learning} (bottom) dataset. Static parts are colored grey while mobile parts are colored green. We also visualize the estimated joint with the red arrow.}
    \label{fig:qual}
    \vspace{-2mm}
\end{figure*}

\subsection{Baselines}

\noindent \textbf{A-SDF.}
    A-SDF~\cite{mu2021sdf} is the closest work to ours, given that no existing method is designed specifically for the full-fledged virtual recreation of articulated objects. There are two main differences between A-SDF and our work. First, A-SDF is a category-level model that assumes the same kinematic tree structures of objects in the same category. Second, it estimates the articulation model implicitly rather than explicitly. Accordingly, we train one A-SDF model for each category and evaluate only the geometry reconstruction result on the synthetic dataset.

\vspace{1mm}
\noindent  \textbf{Correspondence.}
    We first train an FCGF~\cite{choy2019fully} feature extractor on the whole dataset. Then we use the extracted features to find point correspondence across the observations before and after the interaction. An articulation model is fitted based on correspondence and using the non-linear least square algorithm. Besides, we use correspondence to compute the moving distance of every point and segment the mobile points with a threshold of 0.02 on this distance. We reconstruct the mesh of segmented points using a ConvONet~\cite{peng2020convolutional} trained for part reconstruction. This baseline has the same output as Ditto.

\vspace{1mm}
\noindent  \textbf{Global Joint.}
    To validate our choice of dense joint representation, we modify our model and use decoders that predict joint parameters from a global feature. Since the pivot point of a revolute joint axis is ambiguous, \ie, it can move along the axis, we adopt the screw-based joint parametrization in ScrewNet~\cite{jain2020screwnet}. We also apply the loss function of ScrewNet to train the model.

\begin{figure}
    \centering
    \includegraphics[width=0.95\linewidth]{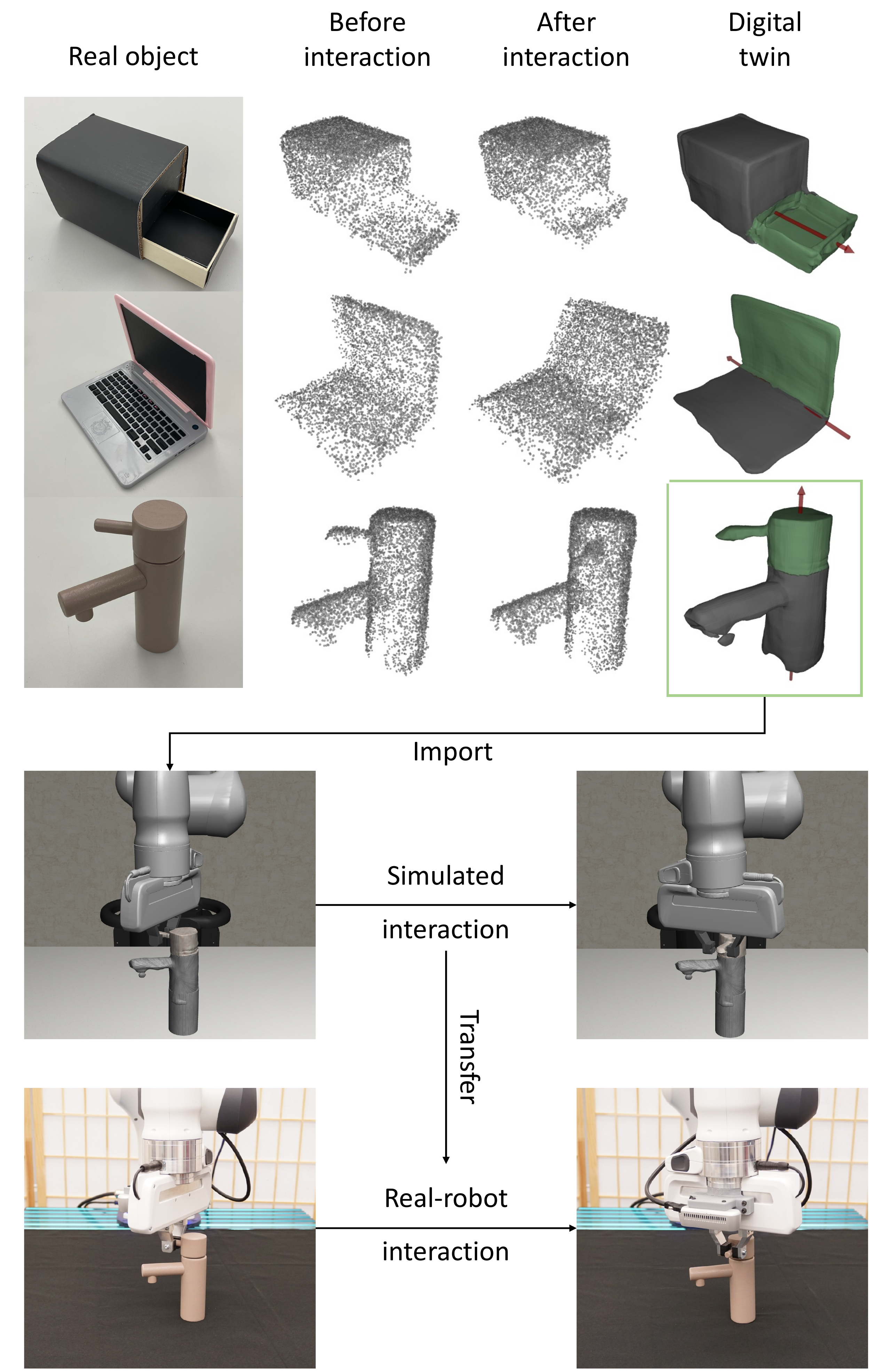}
    \vspace{-2mm}
    \caption{Real-world results. We use Ditto trained in simulated datasets to build the digital twin of these physical objects. The recreated faucet model is imported into a physical simulator. The robot interacts with the virtual faucet and transfers its actions back into the real world to manipulate the physical faucet.}
    \label{fig:real}
    \vspace{-2mm}
\end{figure}

\vspace{2mm}
In addition to the external baselines above, we also use the following ablated versions of our model to validate our design choices:

\vspace{1mm}
\noindent  \textbf{Concat Fusion.}
    Instead of the attention-based fusion, it directly concatenates the structured features of the pair of point clouds and conditions the local implicit decoders on the concatenated features.
    
\vspace{1mm}
\noindent  \textbf{Share Feature.}
    We use 3D feature grids for occupancy prediction and 2D feature planes for segmentation and joint prediction in our current model. This ablated version shares the 3D and 2D features for both geometry and articulation prediction.
    
\vspace{1mm}
\noindent  \textbf{Share Decoder.}
    In our current model, we use two separate decoders in PointNet++ for geometry and articulation. This ablated version uses a shared decoder instead.

\subsection{Evaluation Metrics}
{\flushleft \textbf{Part-level Geometry.}}
    To evaluate the quality of the reconstructed part-level mesh, we use Chamfer-$\ell_1$ distance (CD) as the evaluation metric. Apart from the CD between the whole reconstructed mesh and the ground truth, we also evaluate the CD of the segmented mobile part because it is the only interactable region of the object. CD shown are multiplied by 1000 as in A-SDF~\cite{mu2021sdf}.

{\flushleft \textbf{Articulation Model.}}
    For both types of joints, we measure the axis orientation error (Angle Err). For the revolute joint, we also measure the axis position error (Pos Err) using the minimum distance between the predicted and ground truth rotation axis.

\vspace{-1mm}
\subsection{Articulated Object Reconstruction}
\vspace{-1mm}

The quantitative results are shown in \cref{tab:ablation}. On both datasets, Ditto gets significantly better results on all metrics compared with the baselines. Both the Correspondence~\cite{choy2019fully} and Global Joint~\cite{jain2020screwnet} baselines perform poorly on articulation estimation. %
As shown in \cref{fig:qual}, while the baseline methods produce overall well-reconstructed shapes, the predicted mobile parts have many artifacts. In contrast, Ditto achieves precise part-level geometry reconstruction as well as accurate joint estimation.

Due to the two-stage design, the Correspondence baseline highly relies on a learned disentangled feature representation at the beginning. Bad initial feature representation is prone to inaccurate correspondence and articulation estimation. In comparison, Ditto does not suffer from such a bottleneck as an end-to-end method. The Global Joint baseline performs poorly mainly due to the high variance of direct global joint regression. Failure of joint estimation also harms segmentation prediction because the joint parameter decoders and the segmentation decoder share the same feature planes.
Differently, Ditto predicts joint parameters with respect to each point. The dense predictions are aggregated into the final joint estimation and thus lead to a more robust and accurate result.

To compare with A-SDF~\cite{mu2021sdf}, we provide the shape reconstruction results on the synthetic dataset. When it comes to the whole Chamfer distance, Ditto surpasses A-SDF by a notably large margin. As visualized in \cref{fig:qual}, A-SDF fails to reconstruct the shape details of unseen objects, especially the objects with prismatic joints. In contrast, Ditto accurately reconstructs the whole object and the fine-grained geometry detail like the handles of the cabinet door and drawer. A key difference between A-SDF and ours is that we use a feedforward model while A-SDF uses test-time optimization to find the articulation and shape codes. The inferior performance of A-SDF is due to the interference of articulation code and shape code in test-time optimization. Note that A-SDF requires separate training for each category while Ditto is a category-agnostic method. Thus, the task should be more challenging for Ditto. Furthermore, Ditto can extract explicit articulation and part-level geometry while A-SDF encodes the articulation model implicitly. More comparison with A-SDF is presented in the appendix.

\subsection{Ablation Studies}
\label{sec:ablation}

As shown in \cref{tab:ablation}, Ditto achieves superior or at least on-par performance on all metrics. The mobile Chamfer distance (CD) of Ditto is substantially lower (better) than the ablated versions. Mobile CD measures the quality of the reconstructed mobile part, which is vital for simulating interaction. Share Feature baseline has the worst performance in Mobile CD. We observe that using the same 3D and 2D features for geometry and articulation makes training unstable, and 2D features would harm the reconstruction due to the loss of spatial information after projection. Concat Fusion does not reason about correspondence explicitly and thus shows inferior performance on all metrics compared with Ditto. Finally, the Share Decoder baseline applies one decoder for both geometry and motion features. This decoder needs to reason about geometry and articulation simultaneously. Suffering from a limited capacity, this baseline obtains sub-optimal performance on Mobile-CD and joint angle errors. Qualitative results and analysis of ablation study are in the appendix.

\subsection{Real-World Experiments}

Finally, we use Ditto to recreate digital twins of real-world objects. We choose three daily objects, a toy cabinet, a laptop, and a faucet. The results are shown in \cref{fig:real}. The results have some artifacts due to the noisy and incomplete input point clouds from depth cameras. Despite these artifacts, Ditto can generally reconstruct the geometry and articulation of these physical objects. Moreover, we import the digital twin of the faucet into Robosuite~\cite{robosuite2020}, a robot learning simulation framework. We use a simulated robot arm to interact with the digital twin and transfer the actions back to the real world after calibrating the simulated and real robot frames. The video of this experiment is provided on the project website. With Ditto we can recreate a real-world articulated object to the digital twin in a virtual environment and map the interactions with the digital twin back to actions in the real world.

\subsection{Limitations}
\label{sec:limit}

{\flushleft \textbf{Kinematic tree.}}
Currently Ditto only segments the object into two parts, the mobile and static ones. We hope to extend our method to reconstruct the full kinematic tree of a composite object with multiple joints and parts via consecutive interactions and aggregation of model inference after each interaction. 

\vspace{-2mm}
{\flushleft \textbf{Active perception.}}
We use interactions to create novel sensory data for inferring articulation. These interactions are either by setting the joint states (in simulation) or by human (real world). We hope to develop algorithms that enable an agent to autonomously interact with objects to actively collect data.

\section{Conclusion}
\label{sec:summary}

We introduce Ditto, an implicit neural representation-based model for recreating digital twins of articulated objects through interactive perception. Ditto is an end-to-end model that jointly learns full-fledged geometry reconstruction and articulation estimation from two visual inputs before and after articulated motions. %
Results show that Ditto achieves significantly more accurate results on geometry and articulation reasoning over baselines. Furthermore, we demonstrate Ditto generalizes to real-world objects, and we can directly spawn the recreated digital twins in interactive simulation. These results manifest the potential of autonomous digital twin building in empowering embodied AI research and AR/VR applications.

\subsection*{Acknowledgments}
\noindent We would like to thank Yifeng Zhu for his help with real robot experiments. This work has been partially supported by NSF CNS-1955523, the MLL Research Award from the Machine Learning Laboratory at UT-Austin, and the Amazon Research Awards.

{\small
\bibliographystyle{ieee_fullname}
\bibliography{egbib}
}

\clearpage
\newpage
\appendix

\begin{figure}[t]
    \centering
    \includegraphics[width=\linewidth]{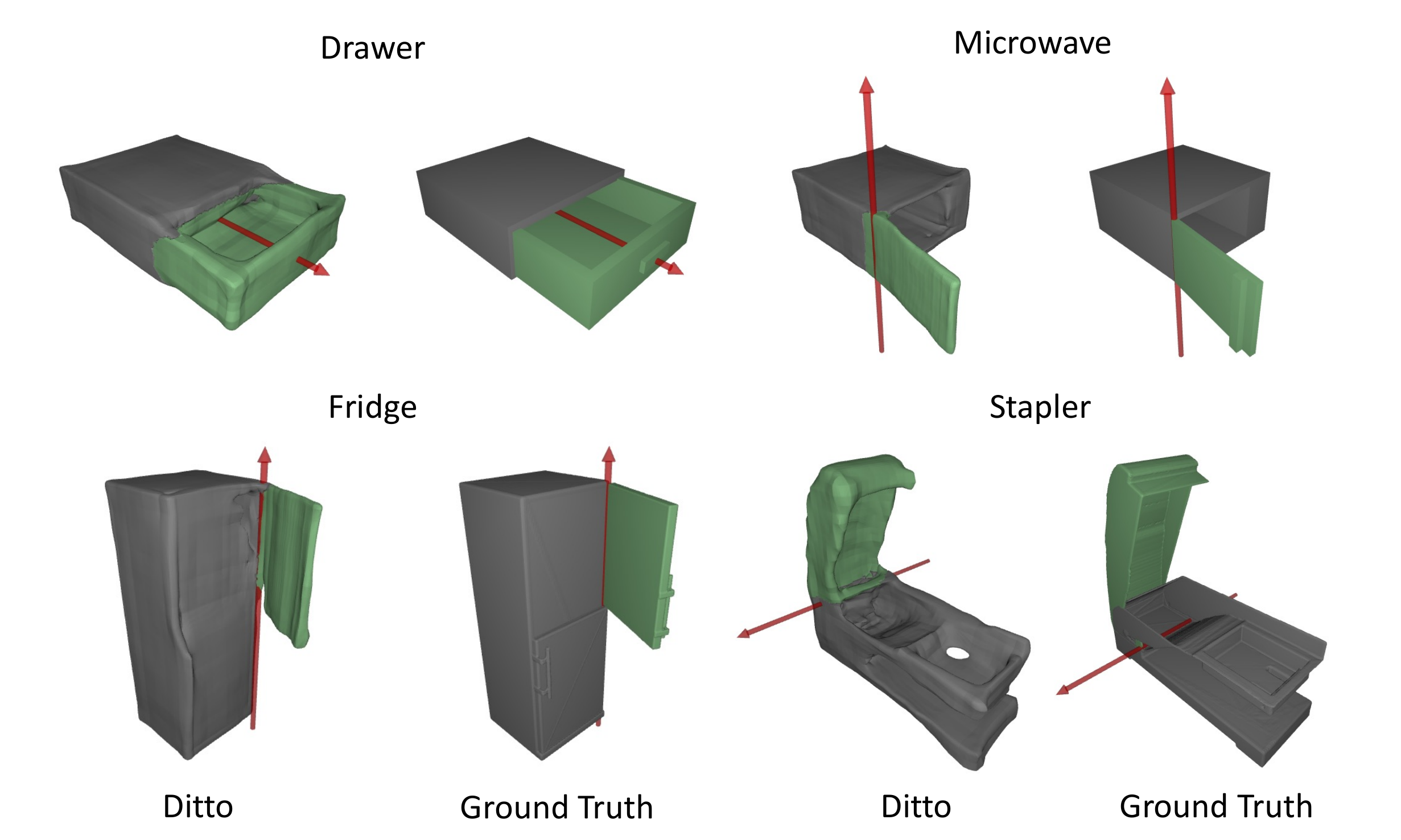}
    \caption{Qualitative results of generalizing to unseen categories.}
    \label{fig:generalize}
\end{figure}

\begin{figure*}[t]
    \centering
    \includegraphics[width=\linewidth]{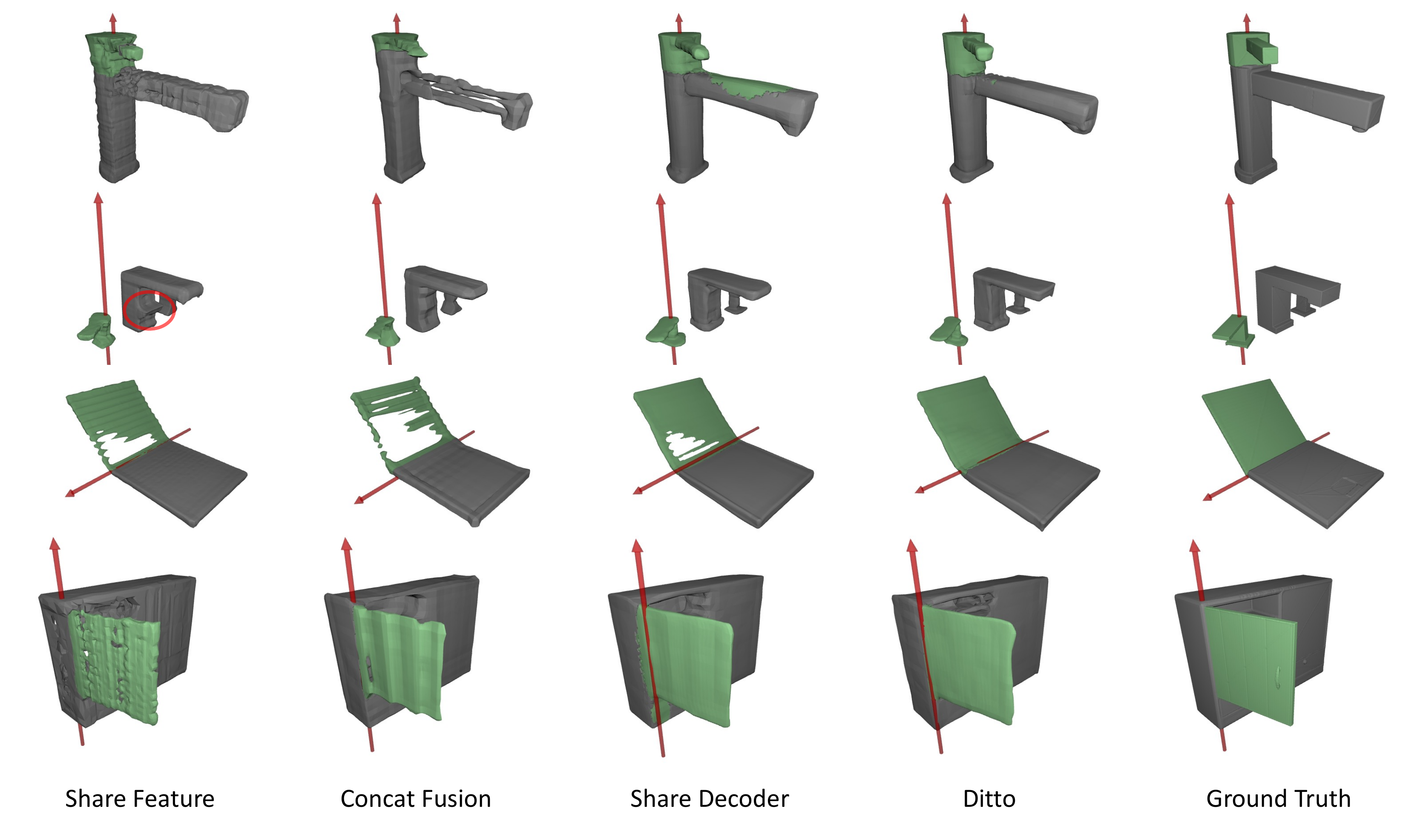}
    \caption{Reconstructed unseen articulated objects in the Shape2Motion~\cite{wang2019shape2motion} dataset of ablated versions. Static parts are colored grey while mobile parts are colored green. We also visualize the estimated joint with the red arrow.}
    \label{fig:qual_ablation}
\end{figure*}

\begin{figure*}[t]
    \centering
    \includegraphics[width=\linewidth]{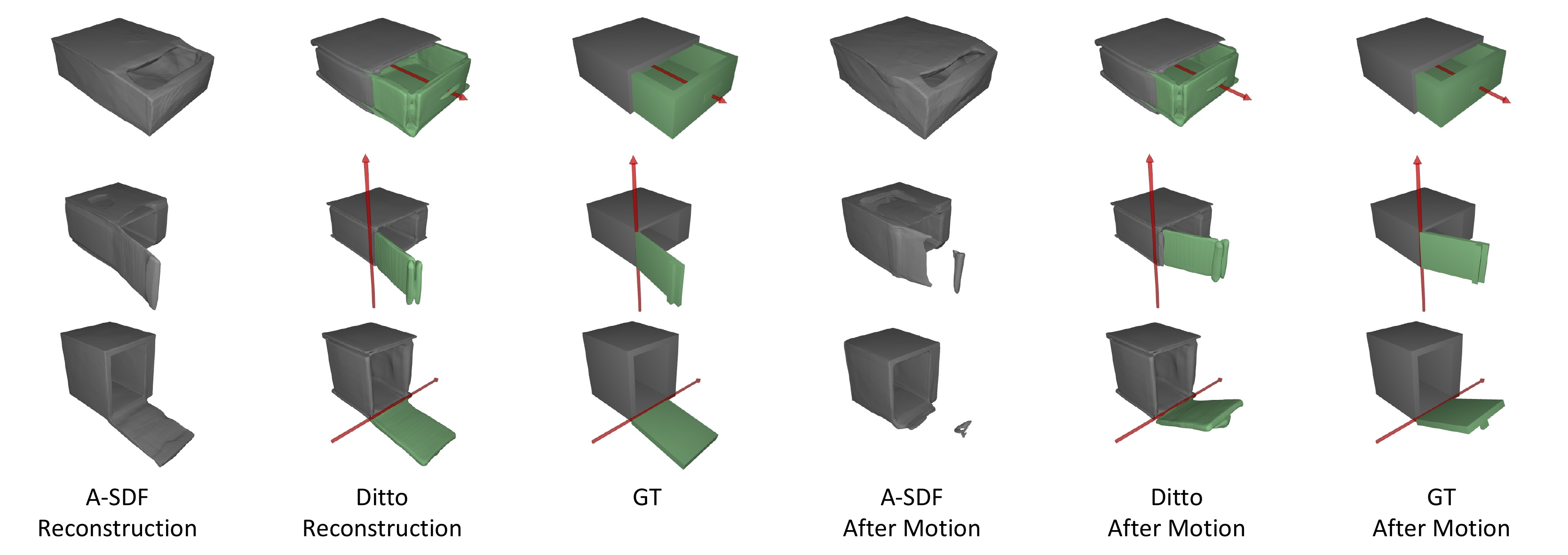}
    \caption{Objects after articulated motion on the Synthetic~\cite{abbatematteo2019learning} dataset. Static parts are colored grey while mobile parts are colored green. We also visualize the estimated joint with the red arrow.}
    \label{fig:qual_motion}
\end{figure*}

\section{Implementation Details}

We use the Shape2Motion dataset~\cite{wang2019shape2motion} and the Synthetic dataset~\cite{abbatematteo2019learning}. The Shape2Motion dataset is licensed under the GNU General Public License v3.0. We sample 8,192 points for each input point cloud. In each iteration, we sample 2,048 pairs of $\mathbf{p}$ and corresponding occupancy. We also 512 pairs of $\mathbf{p}_{in}$ inside the object and corresponding segmentation and joint parameters as query points and ground truths. $\mathbf{p}$ and $\mathbf{p}_{in}$ are input query points for geometry decoder and articulation decoders separately, as described in \cref{sec:decoder}.

We implement the models with Pytorch~\cite{NEURIPS2019_9015} and train the models with the Adam~\cite{kingma2014adam} optimizer and a learning rate of $10^{-4}$ and batch sizes of 8.

\section{Ablation Study}

Ditto uses 3D feature grid for geometry reconstruction and 2D feature plane for articulation estimation. To validate the advantage of this design, we evaluate another ablated version where two 3D feature grid are used for geometry and articulation respectively. As in \cref{tab:oracle}, this ablated version has similar performance to Ditto. But it requires around 20\% more memory usage and training time compared with Ditto.

We show some qualitative results in \cref{fig:qual_ablation}. Our full Ditto model can recreate the articulated objects more accurately, especially the mobile part, benefiting from the attention-based fusion, separate decoders and features. In comparison, Concat Fusion and Share Feature are not able to reconstruct the smooth and complete surface. The ablated version with Share Feature uses 2D feature planes along with 3D feature grids for geometry reconstruction. This projection operation results in artifacts as in the faucet result in \cref{fig:qual_ablation} (red circle in the second row, first column). The ablated version with Share Decoder has a problem segmenting the mobile and the static parts correctly. Overall, Ditto can achieve the best performance on the reconstruction of the articulated objects.

\section{Comparison with A-SDF}

To explore the possible reason behind the inferior performance in \cref{tab:ablation}, we try fixing the A-SDF's articulation code to the ground-truth one. The reconstruction result is improved and close to Ditto as in \cref{tab:oracle}. It indicates that the inferior performance of A-SDF is caused by the in-
terference between articulation and shape codes in test-time optimization. For example, the shape code degrades when the articulation code is in a local minimum far from the ground truth. In contrast, the articulation and geometry predictions do not interfere with each other in our model.

\begin{table}[t]
    \centering
    \begin{tabular}{l|c|c}
    \toprule
    Method & Whole CD $\downarrow$ & Mobile CD \\
    \midrule
     A-SDF (oracle code) ~\cite{mu2021sdf} & 0.66 & - \\
     Ditto (3D+3D feature) & 0.27 & 0.12 \\
     Ditto & 0.25 & 0.16 \\
    \bottomrule
    
    \end{tabular}
    \caption{Quantitative results of reconstruction on the Synthetic~\cite{abbatematteo2019learning} dataset.}
    \label{tab:oracle}
\end{table}

\begin{table}[t]
    \centering
    \begin{tabular}{l|c}
    \toprule
        Method & Chamfer Distance $\downarrow$ \\
        \midrule
        A-SDF~\cite{mu2021sdf} & 3.57 \\
        Ditto & \textbf{0.37} \\
    \bottomrule
    
    \end{tabular}
    \caption{Quantitative results of articulated motion synthesis on the Synthetic~\cite{abbatematteo2019learning} dataset.}
    \label{tab:motion}
\end{table}

\begin{table}[t]
    \centering
    \begin{tabular}{l|c}
    \toprule
    Method & Joint type accuracy (\%)\\
    \midrule
    Global Joint~\cite{choy2019fully} & 88 \\
    Share Feature & 100 \\
    Concat Fusion & 96 \\
    Share Decoder & 100 \\
    Ditto (Ours) & 100 \\
    \bottomrule
    
    \end{tabular}
    \caption{Quantitative results of joint type prediction accuracy on the Shape2Motion~\cite{wang2019shape2motion} dataset.}
    \label{tab:joint_type}
\end{table}

A-SDF~\cite{mu2021sdf} can control the joint state by changing the articulation code. On the other hand, Ditto explicitly reconstructs the explicit part-level meshes and the articulation model, where the joint state can also be easily controlled. It is thus possible to compare the performance of articulated motion synthesis of A-SDF and Ditto. We first reconstruct the articulated object and then manipulate the articulated object to a new joint state. And we measure the whole Chamfer distance between manipulated results and the ground truth objects after such an articulated motion.

The quantitative results are in \cref{tab:motion}. Ditto achieves significantly better results compared with A-SDF. We also show some qualitative results in \cref{fig:qual_motion}. Even though A-SDF can generally reconstruct the articulated object from the observation, the results after the articulated motion are not consistent with the initial state due to its latent representation of articulation. For example, the whole drawer body is widened after the motion. In contrast, Ditto explicitly extracts mobile part mesh and the corresponding joint parameters. Apart from the rigid transformation induced by articulated motion, there is no unexpected distortion after the motion.

\section{Joint Type Prediction}

Our model is also predicting the joint type. All methods give 100\% joint type accuracy on the Synthetic dataset. We provide the results on the accuracy of joint type prediction on the Shape2Motion dataset in \cref{tab:joint_type}. Most methods also acquire 100\% accuracy on this dataset except the global joint baseline and the ablated version with concat fusion.

\section{Generalization to Unseen Categories}

In our experiments, we trained our model for four categories altogether and evaluated it on the same four categories. To test generalization to novel categories, we run our model, trained on Shape2Motion, on four unseen categories (drawer, microwave, fridge, and stapler). \cref{fig:generalize} shows that our model generalizes robustly to geometrically similar categories (drawer, microwave, and fridge) but slightly worse on the new categories of more significant differences (stapler) as it learns shape priors from the training data.

\end{document}